\pdfoutput=1

\documentclass[11pt]{article}

\usepackage{EMNLP2023}

\usepackage{times}
\usepackage{latexsym}

\usepackage[T1]{fontenc}

\usepackage[utf8]{inputenc}

\usepackage{microtype}

\usepackage{inconsolata}

\usepackage{amsmath}
 \usepackage{graphicx}
 \usepackage{multirow}
\usepackage{xcolor}
\usepackage{subcaption}

\definecolor{red}{RGB}{255,0,0} 
\definecolor{blue}{RGB}{0,0,255}

\title{Probing Quantifier Comprehension in Large Language Models: \\Another Example of Inverse Scaling}

\author{Akshat Gupta \\
  AI Research, JPMorgan Chase\thanks{ This work was done while at AI Research, JPMorgan.}\\ University of California at Berkeley\\
  \texttt{akshat.gupta@berkeley.edu} \\}

\begin{document}
\maketitle
\begin{abstract}
With their increasing size, large language models (LLMs) are becoming increasingly good at language understanding tasks. But even with high performance on specific downstream task, LLMs fail at simple linguistic tests for negation or quantifier understanding. Previous work on quantifier understanding in LLMs show inverse scaling in understanding \textit{few}-type quantifiers. In this paper, we question the claims of previous work and show that it is a result of inappropriate testing methodology. We also present alternate methods to measure quantifier comprehension in LLMs and show that LLMs are able to better understand the difference between the meaning of \textit{few}-type and \textit{most}-type quantifiers as their size increases, although they are not particularly good at it. We also observe inverse scaling for \textit{most}-type quantifier understanding, which is contrary to human psycho-linguistic experiments and previous work, where the model's understanding of \textit{most}-type quantifier gets worse as the model size increases. We do this evaluation on models ranging from 125M-175B parameters, which suggests that LLMs do not do as well as expected with quantifiers. We also discuss the possible reasons for this and the relevance of quantifier understanding in evaluating language understanding in LLMs.
\end{abstract}

\section{Introduction}

Large Language Models (LLMs) are getting increasingly better at understanding language \citep{devlin2018bert, radford2019language, raffel2020exploring, zhang2022opt, ouyang2022training, touvron2023llama} which can be seen by their improving performance on various language understanding benchmarks \citep{wang2018glue, wang2019superglue}. Auto-regressive LLMs including encoder-decoder models like BART \cite{lewis2019bart} and T5 \cite{raffel2020exploring} and decoder-only models like GPT \citep{radford2018improving, radford2019language, brown2020language, zhang2022opt, touvron2023llama} have been scaled to billions of parameters to improve their language understanding capabilities. With increasing model sizes, the models also gets increasingly better at learning from context and can just be prompted with few examples rather than fine-tuning to do downstream task \citep{brown2020language, liu2023pre}. 

\begin{table*}
  \centering
\begin{tabular}{c|c|c}
  \textbf{Backbone Phrase} & \textbf{Quantifier} & \textbf{Typicality} \\
  \hline
  
  \multirow{8}{*}{postmen carry} & \multirow{2}{*}{M : \textit{\textcolor{red}{Most}} postmen carry}  & (M, T) : \textit{\textcolor{red}{Most}} postmen carry \textcolor{blue}{mail} \\
  \cline{3-3}
  &  & (M, A) : \textit{\textcolor{red}{Most}} postmen carry \textcolor{blue}{oil} \\
  \cline{2-3}
  & \multirow{2}{*}{F : \textit{\textcolor{red}{Few}} postmen carry}  & (F, T) : \textit{\textcolor{red}{Few}} postmen carry \textcolor{blue}{mail} \\
  \cline{3-3}
    &  & (F, A) : \textit{\textcolor{red}{Few}} postmen carry \textcolor{blue}{oil} \\
  \cline{2-3}
  
 & \multirow{2}{*}{M : \textit{\textcolor{red}{Almost all}} postmen carry}  & (M, T) : \textit{\textcolor{red}{Almost all}} postmen carry \textcolor{blue}{mail} \\
  \cline{3-3}
  &  & (M, A) : \textit{\textcolor{red}{Almost all}} postmen carry \textcolor{blue}{oil} \\
  \cline{2-3}
  & \multirow{2}{*}{F : \textit{\textcolor{red}{Almost no}} postmen carry}  & (F, T) : \textit{\textcolor{red}{Almost no}} postmen carry \textcolor{blue}{mail} \\
  \cline{3-3}
    &  & (F, A) :  \textit{\textcolor{red}{Almost no}} postmen carry \textcolor{blue}{oil} \\

\end{tabular}
  \caption{An example from the dataset used in this paper where a backbone phrase is modified by quantifiers and followed by typical or atypical critical words. }
\end{table*}\label{Table:dataset}

Even with this unprecedented yet implicit evidence of increasing language understanding capability of LLMs, these models still fail simple linguistic tests on understanding negation and quantifiers \citep{jang2023can, kalouli2022negation, michaelov2022rarely}. Understanding negation and quantifiers is challenging for language models because the presence of a single negating or quantifying word can drastically change the meaning of the sentence. Also, such sentences are infrequently used in pre-training text corpora \citep{jimenez2020corpora, michaelov2022rarely}, which makes it hard for the models to account for such situations. Due to this, actual comprehension of \textit{negation} or \textit{quantifier} words is overpowered by the larger context of the sentence, which makes it challenging for language models to deal with these situations. 

We focus on one specific linguistic phenomenon, which is the use of \textit{quantifiers}. Quantifiers are words that usually occur before a noun to express the quantity of an object \cite{kalouli2022negation}. The presence of different quantifiers can make statements semantically very different from each other. It can be seen from the following example:

\begin{align}
(\text{Ex:1})\text{     } & \text{All Ps are Qs} \implies P \subseteq Q \nonumber \\    
&\text{No Ps are Qs} \implies P \cap Q = \emptyset \nonumber
\end{align}\label{eq:example}

In the above example, two different quantifiers \textit{all} and \textit{no} when applied to the sets P and Q end up in polar opposite meanings as can be seen on the right side of respective equations. \textit{All Ps are Qs} means that all objects in the set P belong to the set Q, whereas \textit{No Ps are Qs} means that P and Q are mutually exclusive sets. This minor distinction in the sentence has a drastic effect on the relationship between P and Q. 

In this work, we aim to test and quantify the ability of LLMs to understand quantifiers and how this understanding changes as the models scale. We build upon the work of \cite{michaelov2022rarely}, who test understanding and sensitivity of LLMs for \textit{most}-type and \textit{few}-type quantifiers. They do these tests on a dataset of 960 sentences created using a previously published study on human response (measured using N400 amplitude) to different quantifiers \cite{urbach2010quantifiers}. They find that while LLMs do increasingly well on understanding \textit{most}-type quantifiers, while their understanding of \textit{few}-type quantifiers diminishes as the size of these language models increase. This is an example of an inverse-scaling law \citep{mckenzie2022inverse, wei2022inverse}, where the model gets worse at doing a task as the model size increases. Inverse scaling laws are rare in natural language processing and important to identify, yet they must be cautiously evaluated \cite{wei2022inverse}. 

In this paper, we first show that conclusions about the inverse-scaling of \textit{few}-type quantifier comprehension in LLMs \cite{michaelov2022rarely} need to be revisited because of a possibly faulty methodology, thus leading to a wrong conclusion about inverse-scaling. We discuss the reasons for this in detail later in the paper. We then propose our own method of measuring \textit{quantifier comprehension} in LLMs. We find that LLMs are able to differentiate between sentences that contain \textit{most}-type versus \textit{few}-type quantifiers quite well and this understanding improves as the model size increases. We measure this by quantitatively evaluating if the models react differently for different types of quantifiers. Although, when we evaluate if the model takes into account the meaning of a quantifier, we find that LLMs comprehend \textit{few}-type quantifiers much better than \textit{most}-type quantifiers. We also find that contrary to the results of \cite{michaelov2022rarely}, \textit{most}-type quantifier comprehension gets worse with increasing model size, thus showing an inverse-scaling law in \textit{most}-type quantifier comprehension. In this study, we evaluate a number of different language model families, with models ranging from a size of 125 million parameters to 175 billion parameters, and find that the results are consistent for all LLMs. 





\section{Dataset and Models}\label{sec:dataset}

The models and dataset used in this paper are identical to the ones used in \cite{michaelov2022rarely}. This work uses the log probabilities produced by different language models to calculate a quantity called surprisal, which is introduced later in the paper. We do not make additional API calls or query models. We simply use the log probabilities released by \cite{michaelov2022rarely}, thus mitigating differences due to experimental conditions. This paper aims to provide an alternative way of interpreting the output logits produced by different LLMs compared to \cite{michaelov2022rarely}.

\subsection{Dataset}

We use the same dataset as used by \cite{michaelov2022rarely} which originates from a set of psycholinguistic experiments done on humans \cite{urbach2010quantifiers}. The dataset consists of 120 different backbone phrases, which are modified by two sets of quantifier and completed by a typical and an atypical continuing word. An example can be seen in Table \ref{Table:dataset}. 

The backbone phrase shown in the example is `postmen carry', which is modified by a \textit{most}-type and a \textit{few}-type quantifier. Following \cite{michaelov2022rarely}, in this paper we study the effects of these two quantifiers and how LLMs interpret them. Each backbone phrase is modified by two \textit{most}-type and two \textit{few}-type modifiers. After the quantifiers are used to modify the backbone phrases, if the language model takes into account the meaning of the word, it should be more likely to produce a word with appropriate typicality. Words that are more typically associated with the backbone phrase are labelled \textit{typical (T)}. For examples, the phrase \textit{"postmen carry"} is typically followed by the word \textit{mail} and not by the \textit{atypical (A)} word \textit{oil}. \textbf{We expect the language model to take into account the quantifier when assigning probabilities to the word following the quantifier-modified phrase}. Each backbone phrase modified by a quantifier is tested to be followed by a \textit{typical} and an \textit{atypical} word. The \textit{typical/atypical} words are also together referred to as \textbf{critical words} in this paper. 

The dataset contains a total of 960 sentences, with 120 unique backbone phrases, with 8 modifications to each sentence as shown in Table \ref{Table:dataset}. We have 2 different quantifier types and two quantifiers per quantifier type, thus making four versions of each backbone phrase. Each quantifier-modified backbone phrase is followed by a typical and atypical word, thus making 8 sentences per backbone phrase.

These sentences were used to measure human brain response to critical words in association with the quantifier used \cite{urbach2010quantifiers}. It was found that humans brain signals produce a spike when an atypical critical word is used with the \textit{most}-type quantifier. This spike in brain activation (called N400 signals) are associated with unexpected events. Hence, these N400 spikes show that the atypical critical words when following a \textit{most}-type quantifier were unexpected/incorrect. A lower activation is seen when the \textit{most}-type quantifier is followed by a typical critical word. This spike in the N400 signal can be explained by a quantity called \textit{surprisal}, which is the negative log-probability of the occurence of a word in that context. This means the less likely the word, the higher the surprisal. It was shown in \cite{michaelov2020well} that surprisal as measured in language models explain these N400 spikes very well, and that GPT-3 is the best single predictor of these N400 spikes in humans \cite{michaelov2023strong}.

\subsection{Models}
To evaluate quantifier comprehension in LLMs, we use five family of models. We use the GPT2 model family (125M-1.5B parameters) \cite{radford2019language}, ElutherAI's GPT models (GPT-Neo 125M, GPT-Neo 1.3B, GPT-Neo 2.7B and GPT-J 6B) \cite{black2022gpt}, the OPT model family (125M - 13B parameters), the GPT-3 model family (2B-175B parameters) and the InstructGPT model family \cite{ouyang2022training} called GPT3.5 in the rest of the paper (2B-175B parameters).

\section{Quantifier Comprehension in LLMs}

In this section, we first present how \cite{michaelov2022rarely} measure quantifier comprehension in LLMs. Specifically, we present two ideas of \textit{surprisal} and \textit{quantifier accuracy} and ways to measure both properties as proposed by \cite{michaelov2022rarely}. Alongside, we also highlight shortcomings of these quantifier comprehension evaluation methods.


\subsection{Surprisal}
As defined in section \ref{sec:dataset}, \textit{surprisal} is the negative log-probability of occurrence of a word given a context, as show below:

\begin{equation}
    \text{S}_p\text{(}w_i\text{)} = - \text{log } p(w_i|w_1,\ldots, w_{i-1})
\end{equation}

where $w_i$ is the critical word under observation and $w_1,\ldots, w_{i-1}$ are the words preceding the critical word in a sentence. The underscore \textit{p} in the surprisal represents that this is the definition of surprisal in prior work. \cite{michaelov2022rarely} acknowledge that words in language models are usually split into subwords. For scenarios when this happens for a critical word, \cite{michaelov2022rarely} suggest to sum up the suprisals of each individual subwords. This essentially means multiplying the probabilities of each subword that makes up the critical word. The use of this definition of surprisal is suboptimal as it does not take into account the effects of subword tokenization. 

Previous work has shown that just summing up subword probability results in skewing of probability values towards words with shorter length, which is why these quantities are normalized by length \cite{brown2020language}. In our setting, this means the critical words split into larger number of subwords is likely to be assigned lower probability and thus higher suprisal than critical words that are split into fewer or no subwords. To normalize the effect of subword length, we propose normalizing the surprisal values by the subword length of the critical word, depicted by $N$, following previous works \cite{brown2020language}. Thus, we define surprisal as shown below:

\begin{equation}
    \text{S(}w_i\text{)} = -  \frac{1}{N} \sum_{\forall v_i \text{ } \in \text{ } \{w_i\} }\text{log } p(v_i|w_1,\ldots, w_{i-1})
\end{equation}

where $w_i$ is the critical word split into a set of $N$-subwords represented by the set $\{w_i\}$ and $v_i$ is a subword that belongs to that set. Surprisal can be understood as a term representing the inverse-probability of occuring of a word in a context. If a word has high probablity of occuring in a context, it will have low surprisal, whereas if a word has a low probablity of occuring in a context, it will have high surprisal. In this work, we will use our definition of surprisal.

\subsection{Quantifier Accuracy}
\cite{michaelov2022rarely} define quantifier accuracy based on the surprisal values for the critical word following a quantifier type. The quantifier accuracy test was motivated by the human brain response experiments done in \cite{urbach2010quantifiers}. The aim of defining quantifier accuracy was to measure if language models take into account the meaning of quantifier words when creating the probability distribution over for the critical word. \cite{michaelov2022rarely} proposes that if LLMs take into account the meaning of quantifiers in a sentence, then the typical critical words will be predicted with larger probability and thus lower surprisal values following a \textit{most}-type quantifier, and the atypical critical word will be predicted with larger probability and thus lower surprisal value with a \textit{few}-type quantifier . 

To illustrate this, we refer to the examples shown in Table \ref{Table:dataset}. For the backbone prompt modified by a \textit{most}-type quantifier - "\textit{Most} postmen carry", an LLM is consider accurate if surprisal for the word \textit{oil} is more than surprisal of the word \textit{mail}, or in other words, $p(\text{mail }| \text{ Most postmen carry}) > p(\text{oil }|\text{ Most postmen carry})$. To succinctly express this, a sentence in the dataset is considered to be \textit{most}-type accurate if for a \textit{\textbf{\underline{m}}ost}-type quantifier modified \textbf{\underline{b}}ackbone \textbf{\underline{p}}hrase (MBP),

\begin{align}
    \scalebox{1.1}{S}(typ|MBP) < \scalebox{1.1}{S}(atyp|MBP) \label{eq:prev_1}
\end{align}

Similarly, for a backbone prompt modified by a \textit{few}-type quantifier - "\textit{Few} postmen carry", an LLM is considered accurate if $p(\text{oil }|\text{ Few postmen carry}) > p(\text{mail }|\text{ Few postmen carry})$. This means that the atypical word is more likely to occur with the \textit{few}-type quantifier. Thus, a sentence is considered to be \textit{few}-type accurate for a \textit{\textbf{\underline{f}}ew}-type quantifier modified \textbf{\underline{b}}ackbone \textbf{\underline{p}}hrase (FBP) if for that phrase, 

\begin{align}
    \scalebox{1.1}{S}(atyp|FBP) > \scalebox{1.1}{S}(typ|FBP) \label{eq:prev_2}
\end{align}

\begin{figure}[t]
  \centering
  \begin{subfigure}[b]{0.4\textwidth}
    \includegraphics[width=\textwidth]{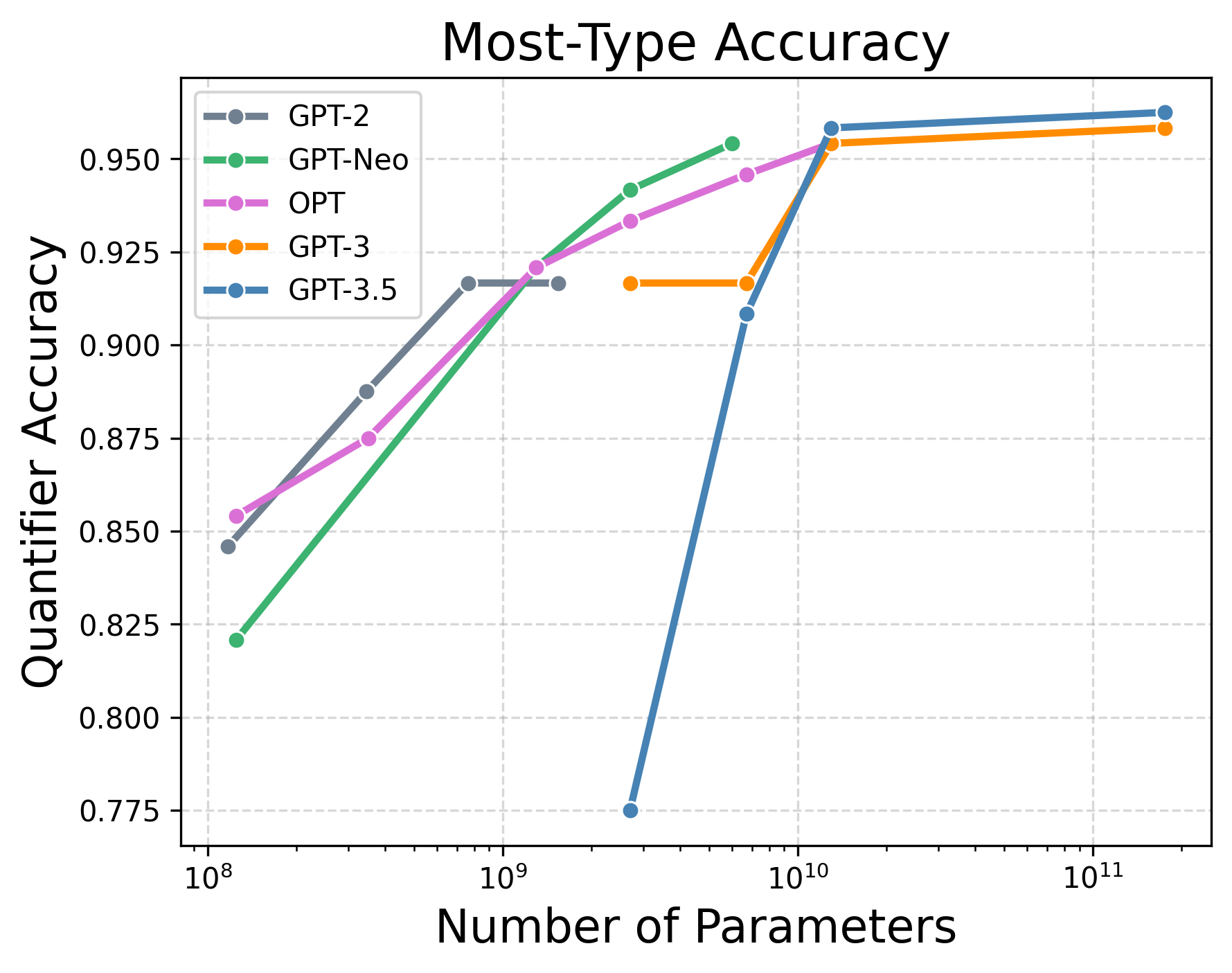}
    \caption{\textit{Most}-type accuracy as measured by \cite{michaelov2022rarely} using equation \ref{eq:prev_1}.\\ }
    \label{fig:c_image1}
  \end{subfigure}
  \hfill
  \begin{subfigure}[b]{0.4\textwidth}
    \includegraphics[width=\textwidth]{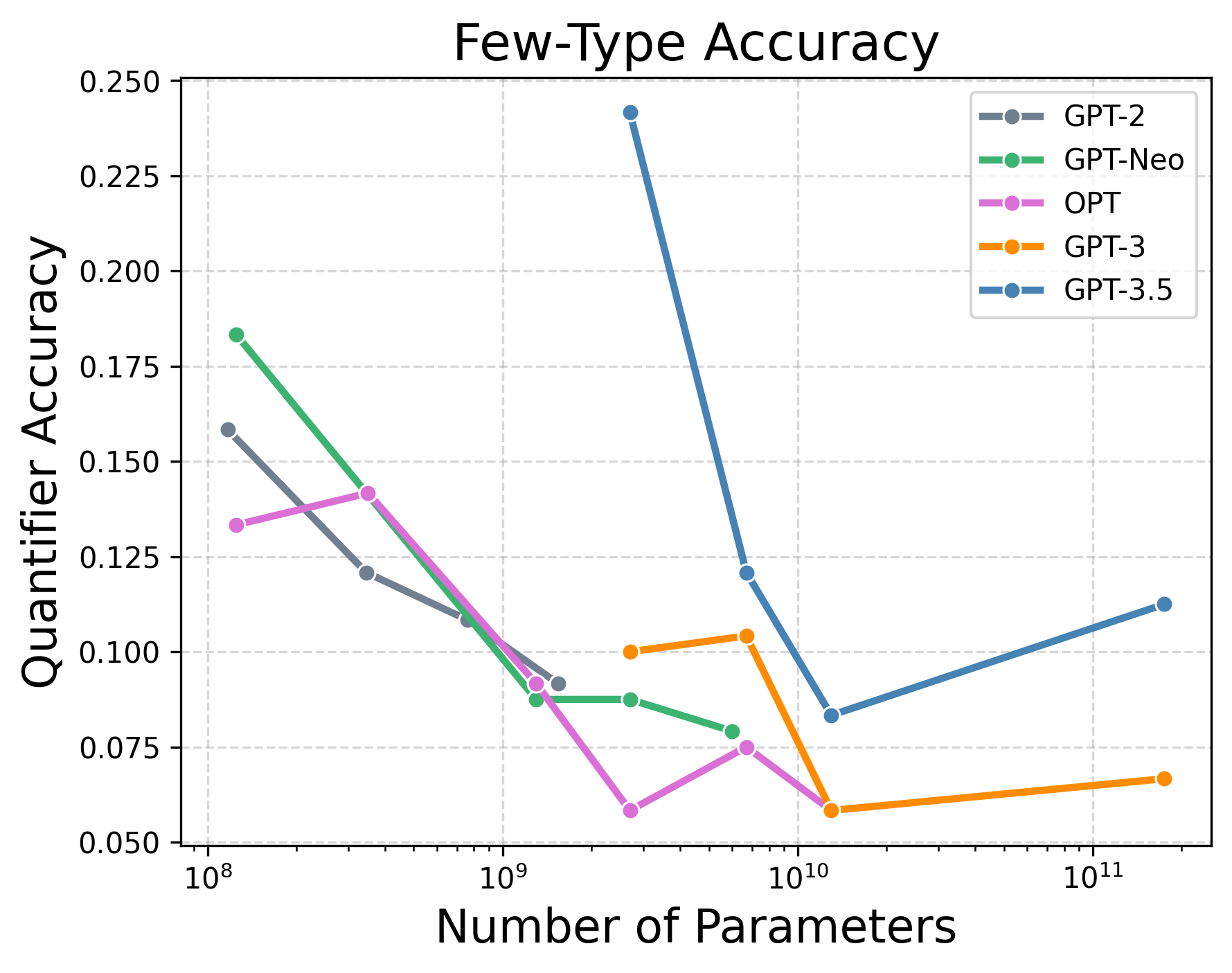}
    \caption{\textit{Few}-type accuracy as measured by \cite{michaelov2022rarely} using equation \ref{eq:prev_1}.\\}
    \label{fig:c_image2}
  \end{subfigure}
  \caption{Quantifier accuracy as a function of model parameters for different models as defined in \cite{michaelov2022rarely}.}
  \label{fig:comparison}
\end{figure}

As proposed by \cite{michaelov2022rarely}, the \textit{most}-type and \textit{few}-type quantifier accuracy is then calculated as the ratio of sentences following the above equations for the respective quantifiers. Figure \ref{fig:comparison} shows \textit{most}-type and \textit{few}-type accuracy for different LLMs as a function of the number of parameters in the model. We also see the inverse-scaling of \textit{few}-type quantifier understanding very clearly. As shown by the plot, as the number of parameters increase, the \textit{few}-type quantifier comprehension gets worse. Figure \ref{fig:comparison} is created using our normalized definition of surprisal taking into account the subword tokenization, and is thus slightly different from the original paper.

\subsubsection{What's wrong with this way of defining quantifier accuracy?}
 Quantifier accuracy as defined in equations \ref{eq:prev_1} and \ref{eq:prev_2} have a few drawbacks. The first is the assumption that \textit{typicality} of a word for humans is the same as that for language models. A word deemed "typical" for a backbone phrase would indeed be in the top few words used by a human, but the same might not be true for language models. To experimentally confirm this, we analyse the output distribution of generated words following a backbone phrase. We find that the "typical" word in the dataset does not even fall into the top-100 most likely words following a backbone phrase for gpt-2 large. This is true for ALL of the sentences in the dataset. \textbf{This shows that the typical token for humans is not necessarily typical for language models}.

\begin{figure}[t]
  \centering
  \begin{subfigure}[b]{0.4\textwidth}
    \includegraphics[width=\textwidth]{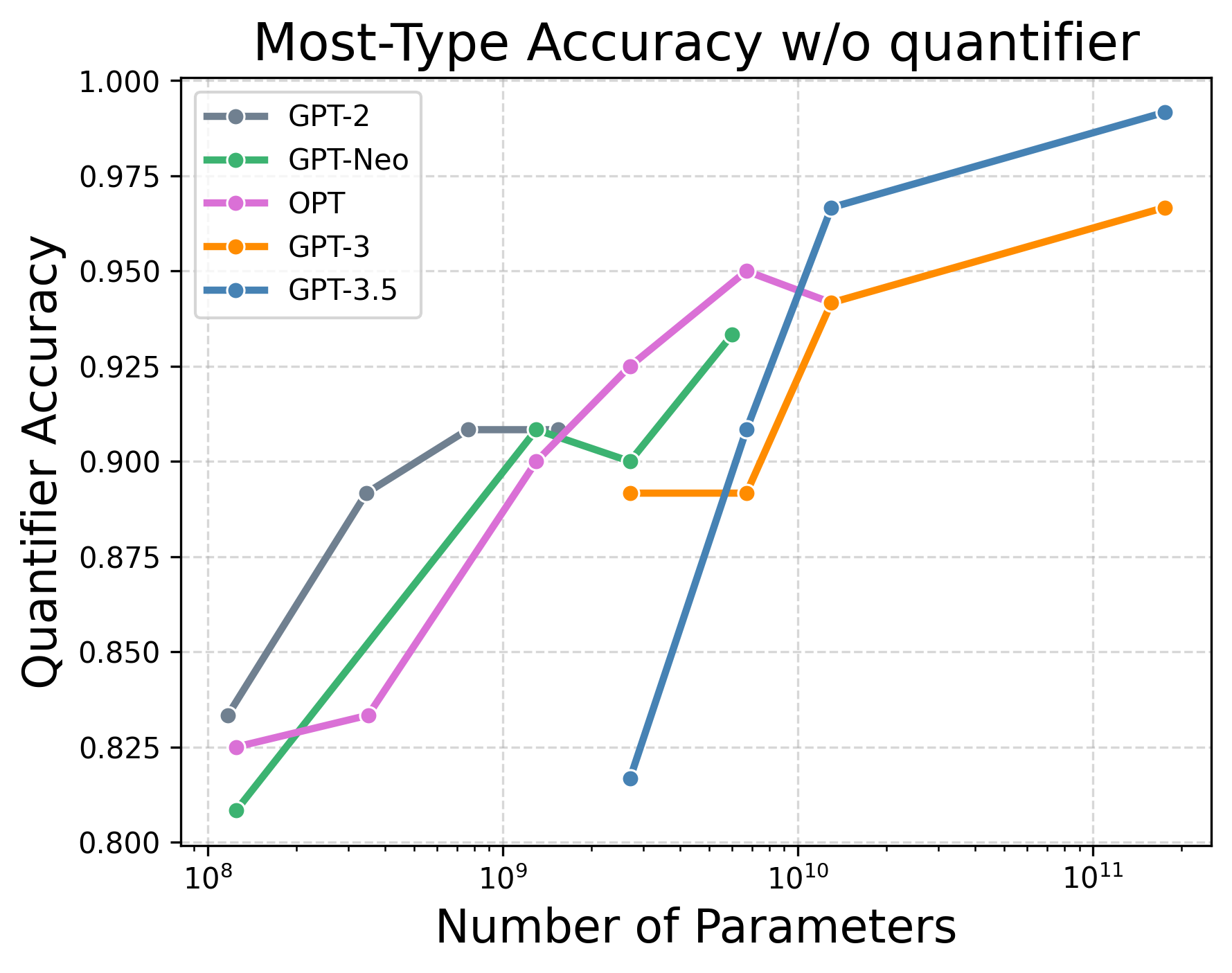}
    \caption{We calculate the \textit{Most}-type accuracy without the quantifier in the context. This just means that we calculate the number of examples where $\scalebox{1.1}{S}(typ|BP) < \scalebox{1.1}{S}(atyp|BP)$. In other words, how often is the typical word followed by the backbone phrase. Note that the modifier is not present in the context here. }
    \label{fig:cp_image1}
  \end{subfigure}
  \hfill
  \begin{subfigure}[b]{0.4\textwidth}
    \includegraphics[width=\textwidth]{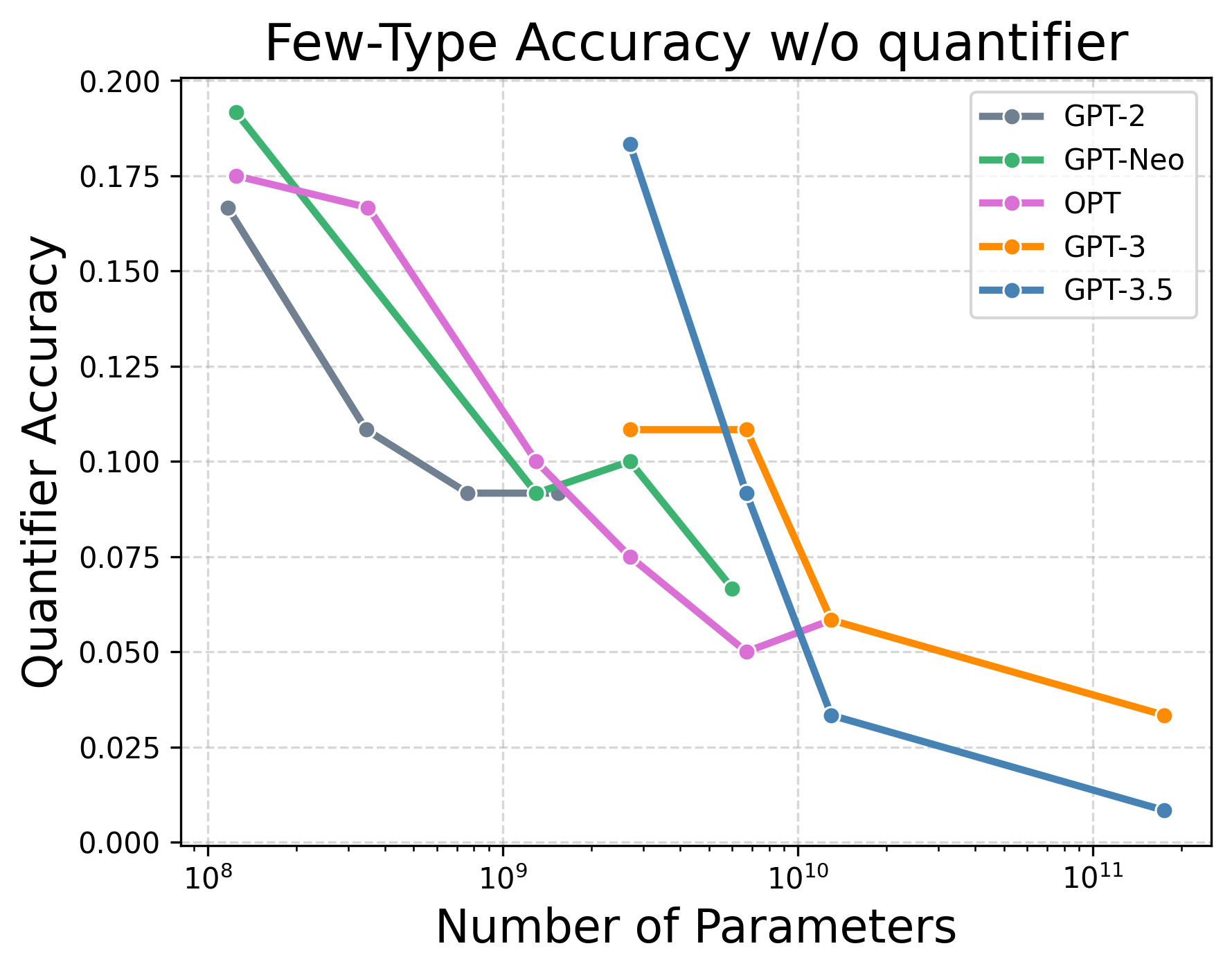}
    \caption{We calculate the \textit{Few}-type accuracy without the quantifier in the context. This just means that we calculate the number of examples where the atypical word is \textbf{not} followed by the backbone phrase, or $\scalebox{1.1}{S}(atyp|BP) > \scalebox{1.1}{S}(typ|BP)$.}
    \label{fig:cp_image2}
  \end{subfigure}
  \caption{Here we calculate the percentage of times the typical words occurs with larger probability than the atypical word in Figure \ref{fig:cp_image1} and vice versa in Figure \ref{fig:cp_image2}. These are similar to the quantities calculated in Figure \ref{fig:comparison} without the quantifier present in the context.}
  \label{fig:comparison_plain}
\end{figure}

 The second assumption is that the chosen atypical word in the dataset is \textbf{the only} complementary word corresponding to the typical word. While the "typical" word is the most common follow up word for a given backbone phrase, we can have many alternative "atypical" words to follow the backbone phrase. For example, if we consider the phrase - "Most postmen carry ", the atypical word \textit{oil} is just as atypical as the word \textit{fish}. In fact, for GPT2-large, \textit{fish} has a larger surprisal value compared to \textit{oil}, which means according to GPT2-large, \textit{fish} is more atypical than \textit{oil} and is thus a more ideal candidate as an "atypical" word for comparison in equations \ref{eq:prev_1} and \ref{eq:prev_2}. Just like the critical word \textit{fish}, we can find many atypical words that are just as atypical if not more, than the chosen words in the dataset. \textbf{This means that if the given atypical word does not satisfy the equations \ref{eq:prev_1} and \ref{eq:prev_2}, there might still exist an unknown number of other atypical words that might be able to satisfy this criteria}. These reasons renders the accuracy metric as defined by \cite{michaelov2022rarely} incorrect.



\subsubsection{What do these scaling graphs actually measure?}

Finally, we want to explain what the scaling in Figure \ref{fig:comparison} and \cite{michaelov2022rarely} actually depicts. To see this, we want to refer the reader to Figure \ref{fig:comparison_plain}, which shows the accuracy metric as defined in equations \ref{eq:prev_1} and \ref{eq:prev_2} for a critical word following a backbone phrase \underline{without} the quantifier. This means that Figure \ref{fig:cp_image1} measures the count when $\scalebox{1.1}{S}(typ|BP) < \scalebox{1.1}{S}(atyp|BP)$, or how often is the typical word followed by the backbone phrase. Similarly, figure \ref{fig:cp_image2} measures $\scalebox{1.1}{S}(atyp|BP) > \scalebox{1.1}{S}(typ|BP)$, or how often the atypical word is \textbf{not} followed by the backbone phrase.

The scaling in Figure \ref{fig:comparison_plain} looks almost identical to Figure \ref{fig:comparison}. This indicates that the method defined by \cite{michaelov2022rarely} to measure the effect of quantifier is not even accounting for the presence of the quantifier, and \textbf{we end up just measuring how often the typical word is more probable than the atypical word}. Thus, the method proposed to evaluate quantifier comprehension using equation \ref{eq:prev_1} and \ref{eq:prev_2} in \cite{michaelov2022rarely} is not actually measuring quantifier comprehension, \textbf{it is measuring typicality}. 



In fact, what these scaling plots show is that as the size of the  model increase, the typical words in LLMs get more probable and the atypical words get less probable. This essentially means that the model is getting better at understanding language as typically used by humans, and is able to associate the typical word in a given context with larger probability than the atypical words.

 \begin{figure}[t]
  \centering
  \begin{subfigure}[b]{0.4\textwidth}
    \includegraphics[width=\textwidth]{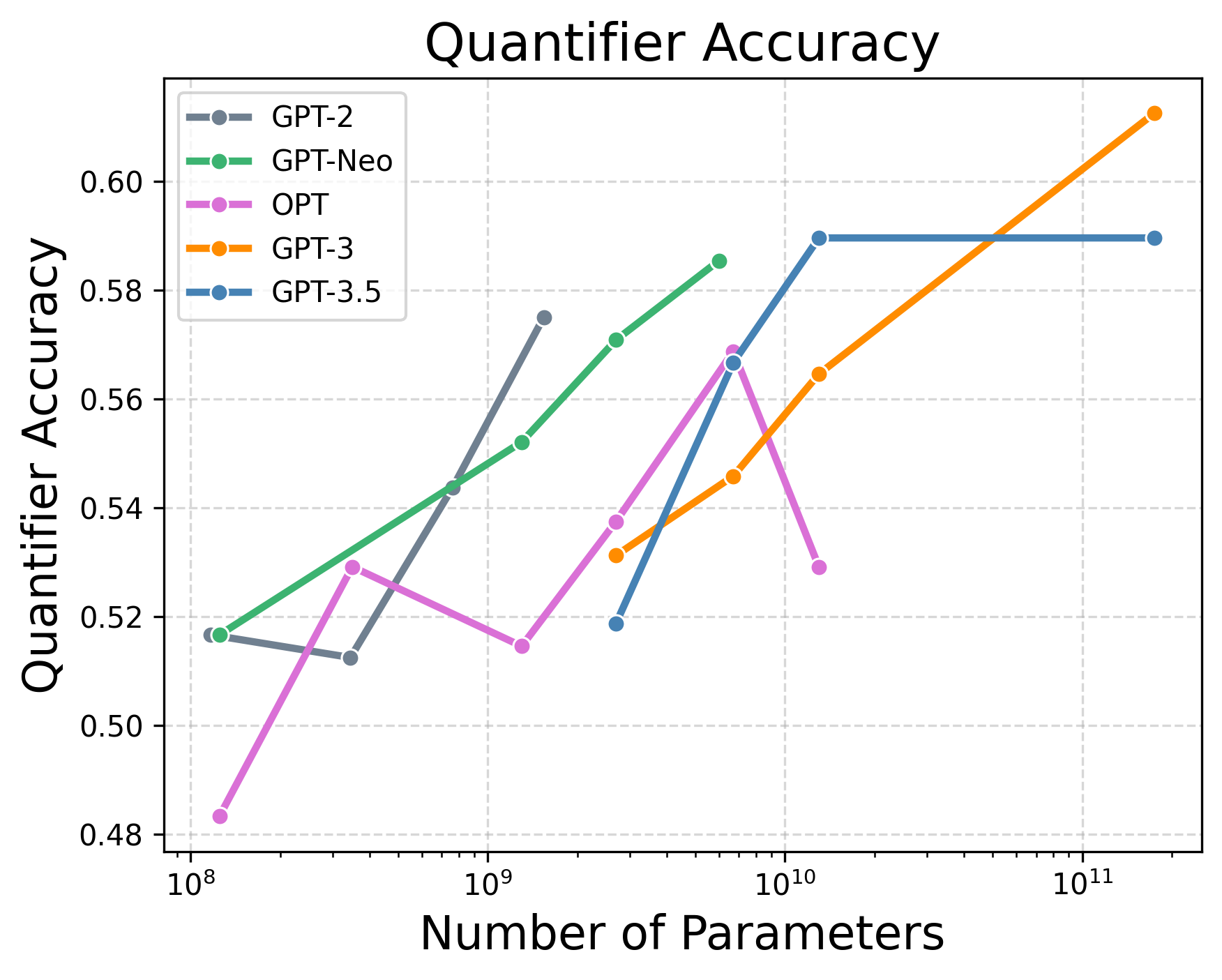}
  \end{subfigure}
  \caption{This figure shows that large language models get increasingly better at differentiating between \textit{most}-type quantifiers and \textit{few}-type quantifiers as they scale.}
  \label{fig:relative}
\end{figure}

\section{Proposed Evaluation of Quantifier Comprehension in LLMs}\label{sec:proposed_definition}

In this section, we present a more robust way of measuring quantifier comprehension in LLMs. Measuring quantifier comprehension in LLMs in the setting defined by \citep{michaelov2022rarely} has to be grounded in the principle that the typical and atypical words chosen in the dataset are not unique, and hence to measure the effect of presence of quantifier in a context, we should do measurements on the same critical word. We propose two tests do this. 

\subsection{EXPERIMENT-1 : Differentiating Between Different Types of Quantifiers}\label{sec:comparison}
In this section, we check if the models are able to differentiate between the meanings of two types of quantifiers and react appropriately. To check this, we fix a critical word (either typical or atypical), and change the quantifier and see how the surprisal value of the critical word is affected. We expect that when we have a typical critical word, the \textit{few}-type quantifier should lead to a higher surprisal value or make the typical word less probable. For example, for the phrase \textit{"Most/Few postmen carry \textcolor{red}{mail}"}, the surprisal for the word \textcolor{red}{mail} should be more when accompanied by a \textit{few}-type quantifier than when compared to a \textit{most}-type quantifier. Similarly, for an atypical word, surprisal values for \textit{most}-type quantifiers should be larger than when observed with \textit{few}-type quantifiers. In summary, an LLM is able to differentiate between two types of quantifiers if for a critical word, one of the following is true depending on the type of critical word under observation:

\begin{align}
    \scalebox{1.1}{S}(typ|MBP) < \scalebox{1.1}{S}(typ|FBP) \label{eq:relative_1}\\
    \scalebox{1.1}{S}(atyp|MBP) > \scalebox{1.1}{S}(atyp|FBP) \label{eq:relative_2}
\end{align}

The results of Experiment-1 are shown in Figure \ref{fig:relative}. We see that LLMs get increasingly better at differentiating between the two types of quantifiers and are able to adapt their output probability distribution at the critical word to reflect this understanding. This improvement of quantifier comprehension scales with increasing model size just like other capabilities of LLMs. Although the absolute value of quantifier accuracy peaks only at about 61\% for the 175 billion parameter GPT-3 model, which shows that for a majority of sentences, the meaning of the quantifier is not reflected in the output probability distribution at the critical word. This shows that although LLMs are getting better at understanding quantifiers as they scale, they are far from perfect.

\subsection{EXPERIMENT-2: Measuring Quantifier-Specific Accuracy}\label{sec:backbone}
Here we want to measure how good LLMs are at understanding a specific quantifier. To measure this, we compare how the surprisal of a critical word is affected as we add a quantifier in the context. When we add \textit{most}-type quantifiers, the surprisal should decrease for a typical word whereas it should increase for an atypical word. In other words, a sentence is accurate for \textit{most}-type quantifier comprehension if:

\begin{align}
    \scalebox{1.1}{S}(typ|MBP) < \scalebox{1.1}{S}(typ|BP) \label{eq:backbone_most_1}\\
    \scalebox{1.1}{S}(atyp|MBP) > \scalebox{1.1}{S}(atyp|BP) \label{eq:backbone_most_2}
\end{align}

Here, MBP is a \textit{\textbf{\underline{m}}ost}-type quantifier modified \textbf{\underline{b}}ackbone \textbf{\underline{p}}hrase, such as \textit{"Most postmen carry"} and BP is just a \textbf{\underline{b}}ackbone \textbf{\underline{p}}hrase without modifier, such as \textit{"Postmen carry"}. Similarly, for \textit{few}-type quantifiers, the surprisal should decrease for atypical critical words and increase for typical words. Specifically, sentence is considered accurate for a \textit{few}-type quantifier comprehension if:

\begin{align}
    \scalebox{1.1}{S}(typ|FBP) > \scalebox{1.1}{S}(typ|BP) \label{eq:backbone_few_1}\\
    \scalebox{1.1}{S}(atyp|FBP) < \scalebox{1.1}{S}(atyp|BP) \label{eq:backbone_few_2}
\end{align}

\begin{figure}[t]
  \centering
  \begin{subfigure}[b]{0.4\textwidth}
    \includegraphics[width=\textwidth]{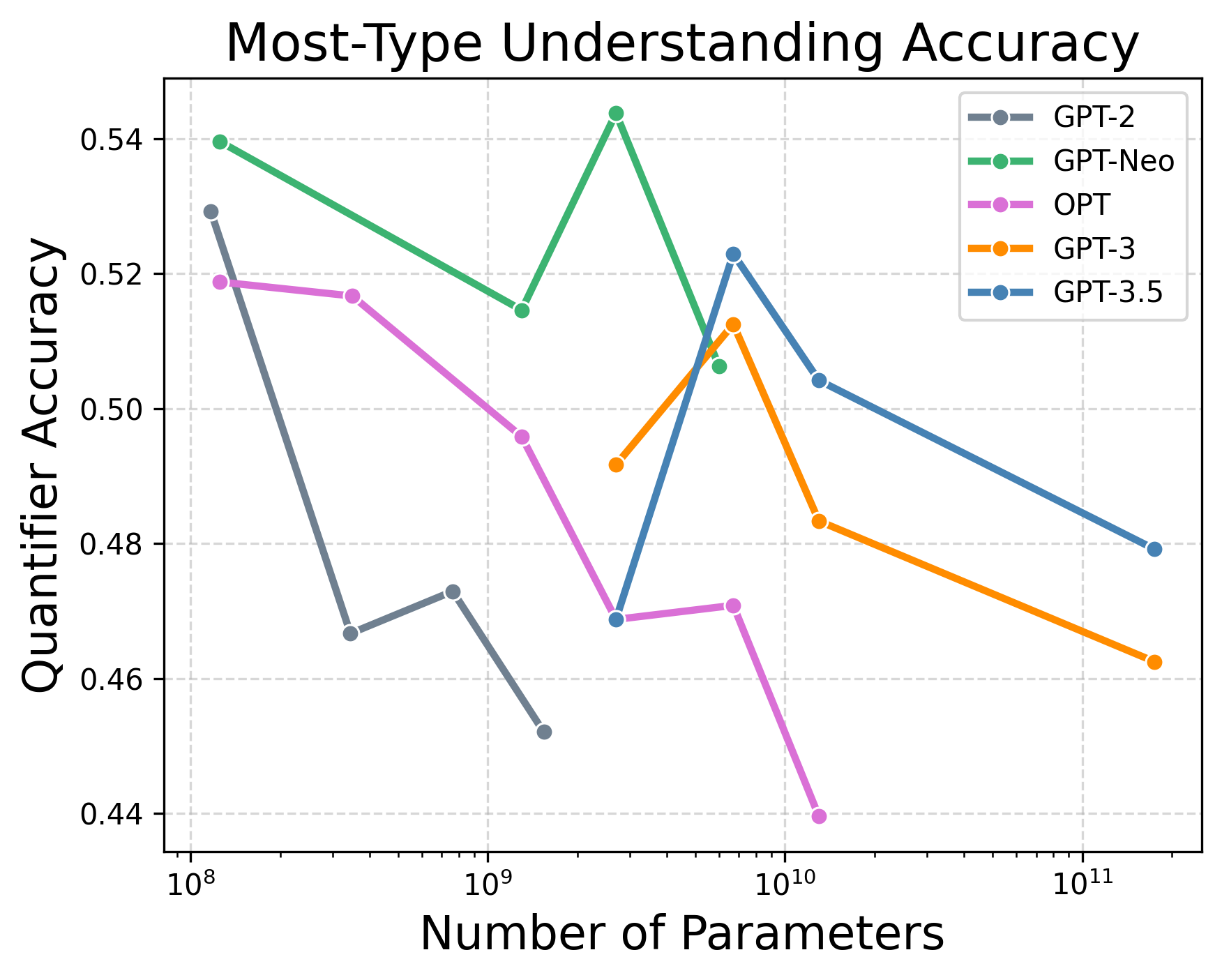}
    \caption{\textit{Most}-type accuracy as defined in equations \ref{eq:backbone_most_1}-\ref{eq:backbone_most_2}\\}
    \label{fig:bb_image1}
  \end{subfigure}
  \hfill
  \begin{subfigure}[b]{0.4\textwidth}
    \includegraphics[width=\textwidth]{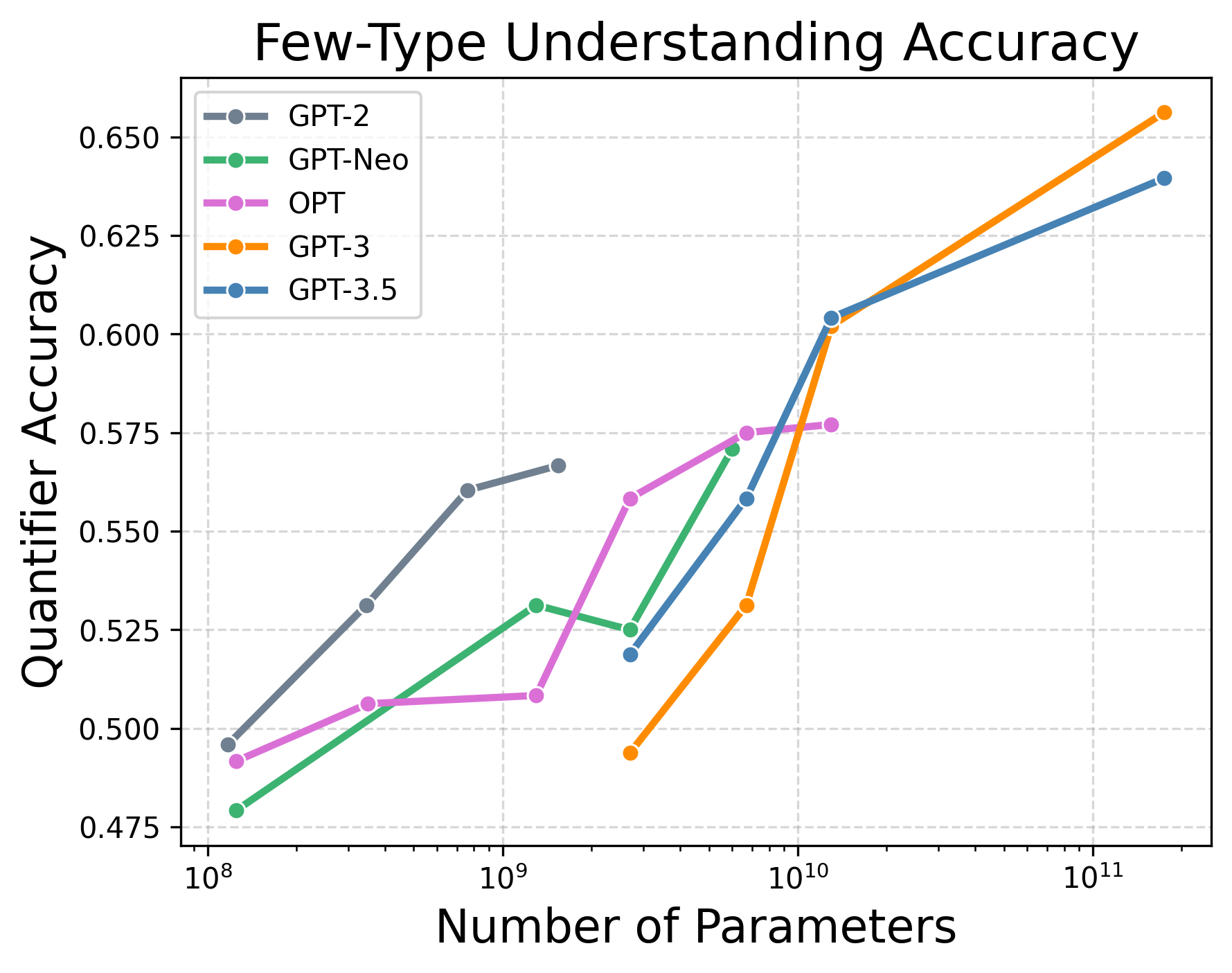}
    \caption{\textit{Few}-type accuracy as defined in equations \ref{eq:backbone_few_1}-\ref{eq:backbone_few_2} \\}
    \label{fig:bb_image2}
  \end{subfigure}
  \caption{Quantifier specific accuracy as defined in equations \ref{eq:backbone_most_1}-\ref{eq:backbone_few_2}.}
  \label{fig:backbone}
\end{figure}

Figure \ref{fig:backbone} shows the quantifier-specific comprehension ability of models as defined in equations \ref{eq:backbone_most_1}-\ref{eq:backbone_few_2}. Although section \ref{sec:comparison} showed that models are able to differentiate between \textit{most}-type and \textit{few}-type quantifiers, we see in Figure \ref{fig:backbone} that they don't necessarily incorporate the meaning of quantifiers when quantifiers are added to a sentence. We see that LLMs become increasingly better at incorporating the meaning \textit{few}-type quantifiers as model size increases by changing the relative probability values of the critical words given the change in context. But this is not observed in the case of \textit{most}-type quantifiers, where we find that the models get increasingly worse at taking into account quantifier meaning, thus showing an \textbf{inverse-scaling in \textit{most}-type quantifier comprehension}. This shows that the model gets increasingly worse at understanding \textit{most}-type quantifier as the size of the model increases. 

Note that in this work, to calculate suprisal, we never compare two different critical words as can be seen in equations \ref{eq:relative_1}-\ref{eq:backbone_few_2}. This circumvents any affects due to subword tokenization and the non-uniqueness of the chosen critical words in the dataset. All the comparisons are made with respect to a single critical word. 


\section{Discussion}
The above two tests for evaluating quantifier comprehension in LLMs show that these models are far from perfect. The underlying premise of the method used in this paper and \cite{michaelov2022rarely} is that the presence of a quantifier should increase or decrease the probability of a critical word depending on its typicality \cite{michaelov2022rarely}. But both tests described in section \ref{sec:proposed_definition} show that this is not ubiquitously observed. The accuracy numbers for both tests are around 50-60\%, which means that the probability distributions do not incorporate quantifier meaning for a large majority of sentences. A test like this makes a fundamental assumption that understanding of meaning can be measured by studying the relative ranking of tokens in the generated word logit. While this is a fair assumption, we think it is necessary to explicitly point this out

Incorporating quantifier meaning in this way is not a necessary condition for models to perform well, as can be seen by their consistent improvement across different benchmark \cite{wang2018glue, wang2019superglue, brown2020language, touvron2023llama}. Also, it has been shown in previous studies that humans are not that great at quantifier comprehension as well \cite{urbach2010quantifiers}, and continue to have a preference towards the more typical word in a context irrespective of the quantifier. These observations suggest two things. Firstly, that LLMs are not good at quantifier comprehension. Secondly, we also observe this lack of sensitivity to quantifier meaning in humans. This combined with the fact that despite lack of quantifier comprehension, LLMs get increasingly better at language understanding, we can argue that quantifier comprehension is not as necessary of a task in language processing and understanding as we thought it was.

\section{Related Work}
Inverse scaling laws were introduced as a competition \cite{mckenzie2022inverse} to incentivize research towards finding scenarios where language models get worse as their size increases. As the field of NLP moves towards scaling models to larger and larger sizes, it is important to know the scenarios where this scaling becomes detrimental \citep{wei2022inverse, mckenzie2023inverse}. 

As language models get increasingly better, some common linguistic tests that they are put through revolve around understanding negation and quantifiers. Studying the affects of negation has been the subject of focus for many studies \citep{kassner2019negated, kalouli2022negation, ettinger2020bert} for different encoder-based masked language models. These studies find that these language models are not sensitive to negations. Studies on quantifiers \citep{kalouli2022negation} also seem to show similar results for masked language models. \cite{michaelov2022rarely} was the first work to study the quantifier understanding in decoder-based LLMs. 

\section{Conclusion}
In this paper, we conduct a study to evaluate how well large language models understand quantifiers. Specifically, we study two types of quantifiers - \textit{most}-type and \textit{few}-type quantifiers. We present a set of experiments to evaluate quantifier comprehension of large language models and show that these models are able to differentiate between \textit{most}-type and \textit{few}-type quantifiers as they scale. We also show that LLMs struggle incorporate the meaning of \textit{most}-type quantifier comprehension when compared to \textit{few}-type quantifiers. We also show that \textit{most}-type quantifier comprehension demonstrates an inverse-scaling law and their understanding of \textit{most}-type quantifiers get worse as the model size increases. This study indicates that LLMs do not take into account the meaning of quantifiers that strongly, as shown by low accuracy scores in Figures \ref{fig:relative} and \ref{fig:backbone}. Even so, these models get increasingly better at language understanding tasks, thus indicating that quantifier understanding might not be the best test to evaluate language understanding in LLMs.



\section*{Acknowledgements}
This paper was prepared for informational purposes in part by the Artificial Intelligence Research Group of JPMorgan Chase \& Co and its affiliates (“J.P. Morgan”) and is not a product of the Research Department of J.P. Morgan.  J.P. Morgan makes no representation and warranty whatsoever and disclaims all liability, for the completeness, accuracy, or reliability of the information contained herein.  This document is not intended as investment research or investment advice, or a recommendation, offer, or solicitation for the purchase or sale of any security, financial instrument, financial product, or service, or to be used in any way for evaluating the merits of participating in any transaction, and shall not constitute a solicitation under any jurisdiction or to any person if such solicitation under such jurisdiction or to such person would be unlawful. 

© 2023 JPMorgan Chase \& Co. All rights reserved.

\bibliographystyle{acl_natbib}
\bibliography{emnlp2023}




\end{document}